# Tuning CLD Maps

R. Marazzato[1] and A.C. Sparavigna[2]
[1]Dipartimento di Automatica ed Informatica, Politecnico di Torino, Italy
[2]Dipartimento di Fisica, Politecnico di Torino, Italy
*amelia.sparavigna@polito.it*, *roberto.marazzato@polito.it*

**Abstract** *The Coherence Length Diagram and the related maps have been shown to represent a useful tool for image analysis. Setting threshold parameters is one of the most important issues when dealing with such applications, as they affect both the computability, which is outlined by the support map, and the appearance of the coherence length diagram itself and of defect maps. A coupled optimization analysis, returning a range for the basic (saturation) threshold, and a histogram based method, yielding suitable values for a desired map appearance, are proposed for an effective control of the analysis process.*

**1 Introduction**
Analysing images through the Coherence Length Diagram (CLD) [1] returns a characteristic polar diagram (the CLD itself) and three image maps: the Support Map (SMAP), which is a graphical outline of areas of the image contributing to the process, the Defect Map (DMAP), which displays differences between local and overall CLD values, and the Directional Defect Map (DDMAP), an outline of shape differences between the CLD of each point and the average CLD. Computing such images involves choosing the value of three parameters. In this paper we discuss reasonable methods to determine an optimal value for one, and a guiding framework to choose the remaining two threshold. For any naming and notation issue please refer to our previous paper [1]. Let us just observe that coherence length is well-know in physics [2,3]: to the authors' knowledge, the CLD mapping, as previously proposed in [4-6], is the only example of a statistical approach to image processing using the coherence length. It is then a processing different from the commonly used statistical processing methods [7,8].

The first value to investigate in CLD mapping is the saturation threshold $\tau$, which defines an interval around the average brightness value $M_0$, in order to decide when the local directional moving average can be considered as "saturated", so determining the coherence length for a certain point and direction (see Fig. 1). The second threshold to set, $\tau'$, defines the range of local CLD lengths to be considered as "successful", with respect to the mean length for each direction. The value (colour) of each point of the DMAP depends on the number of "successful/defective" directions found over all 32 standard directions. Fig. 2 is a graphical outline of this process. A similar procedure is applied in order to obtain the DDMAP, and the third threshold $\tau''$ is the maximum allowed value for the normalised square difference sum, representing the shape mismatch of each local CLD to the overall diagram (see Fig. 3). Here we discuss the methods to find suitable values for each of such settings: a two factors optimisation allows to compute a range for the saturation threshold, while a defect percentage is created in order to allow the analyst to link the threshold to be set to the desired appearance of the corresponding map.

## 2 Optimizing CLD and SMAP

When the saturation threshold $\tau$ is to be set, the goal is finding a solution which takes into account the behaviour of both CLD and SMAP (Support Map, [1]). A rough analysis of the relationship between the CLD and $\tau$ could start by noticing that very large values of $\tau$, such that any allowed brightness satisfies the saturation condition, cause the diagram to collapse to a point, as no additional contributions are needed for any point, in order to saturate the local directional average value to the mean brightness $M_0$. An example is shown in Fig.4, where the behaviour of CLD is depicted for three different thresholds for the image on the right [9].

As $\tau$ decreases, two phenomena happen:
- the length of local CLD components $l_{0,i}^{\tau}(x,y)$ at point $(x,y)$ in the image frame, when computable, tends to increase, as more contributions are required to reach the edge of the saturation range, and
- for some direction of a certain subset of points the coherence length cannot be computed, as the longer path to be followed is not compatible with the dimensions of the image.

The first one causes the overall CLD to increase its size, while the second one affects both the CLD and the SMAP. In facts, when the length contributed by a point disappears, the average length for that direction changes. The SMAP can be interpreted as the average of 32 domains $D_i^{\tau}$, each of whose pertains to a specific direction $i$, and contains only points contributing to CLD in that direction. As the threshold value decreases, computing the coherence length in a given direction becomes impossible for an increasing number of points, what leads to a smaller measure (cardinality) for the corresponding $D_i$ sets.

The behaviour of both CLD and SMAP can be quantitatively described by means of two indexes. The first one is a simple average of all lengths of the overall CLD:

$$\omega(\tau) = \frac{1}{32} \sum_{i=1}^{32} l_{0,i}^{\tau} \qquad (1)$$

while a function of $\tau$ describing the fraction of image surface corresponding to "successful" points is

$$\Omega(\tau) = \frac{1}{hw} \sum_{(x,y) \in D} \Phi^{\tau}(x,y), \qquad (2)$$

where $h, w$ are the height and the width of the image in pixels and $\Phi(x,y)^{\tau}$ is the support map function. The maximum (meaningful) value for $\tau$ can be written as

$$\tau_{max} = \frac{\max_{D} |b(x,y) - M_0|}{M_0}, \qquad (3)$$

so our goal is determining the optimum value of $\tau$ in the interval $(0, \tau_{max}]$. As we would like to obtain the maximum possible value for both $\Omega(\tau)$ and $\omega(\tau)$, considering that this cannot be achieved with the same threshold level, the better choice is finding an appropriate intermediate value, such that both functions are "good enough". The analysis process which allows to determine such value is based on the discussion of the overall quality factor, given by the unbiased product

$$\Pi(\tau) = (\Omega(\tau) - \Omega_0)(\omega(\tau) - \omega_0), \qquad (4)$$

where

$$\Omega_0 = \min_{\tau \in (0, \tau_{max}]} (\Omega(\tau)) \; ; \; \omega_0 = \min_{\tau \in (0, \tau_{max}]} (\omega(\tau)) \qquad (5)$$

Let us assume a test image as in Fig.5. While $\Omega(\tau)$ is strictly non decreasing in the cited interval (see Fig. 6), the behaviour of $\omega(\tau)$ could be more complex, as described earlier; anyway, a restricted subset of images can be considered, for which $\omega(\tau)$ is strictly non increasing in $(0, \tau_{max}]$:

$$B_\omega = \{b : D \to [0,\ldots,255] : \tau > \tau_1 \Rightarrow \omega(\tau) \le \omega(\tau_1)\}. \qquad (6)$$

Our discussion will only consider such images. From a practical point of view, all tested images showed a behaviour of this type (see Fig.6.a,b). Under these conditions, the function $\Pi(\tau)$ shows a maximum for a certain point $\tau_0 \in (0, \tau_{max})$ (see Fig. 6.c). The value $\tau_0$ is the desired optimal threshold, and can be computed by means of elementary routines [10].

## 3 Setting the success percentage in Defect Maps

A DMAP contains "successful" and "defective" pixels, which are computed through the comparison of the local CLD to the overall image CLD. Depending on the threshold $\tau'$, the amount of "successful" pixels takes different values. The goal of this section is describing a function which links such amount, which can be chosen by the analyst testing the image, to $\tau'$. Let us consider a pixel $(x,y) \in D$, and call

$$\tau'_{max,\tau}(x,y) = min\left(\tau' \in R : l_{0,i}^\tau(x,y) \in \Delta_{\tau,\tau'}^i, \forall i = 1,\ldots,32\right) \qquad (7)$$

the minimum threshold value for which the local coherence length is close enough to the average length for all directions. For the specified pixel, all useful threshold values are less or equal to this bound, so it can be considered the maximum threshold. From a practical point of view, it can be computed as

$$\tau'_{max,\tau}(x,y) = max\left(\frac{|l_{0,i}^\tau(x,y) - l_{0,i}^\tau|}{l_{0,i}^\tau}\right). \qquad (8)$$

Thus, the overall maximum threshold value is

$$\tau'_{max,\tau} = \max_{(x,y) \in D} \left(\tau'_{max,t}(x,y)\right). \qquad (9)$$

Now consider the defect map $\Psi^{\tau,\tau'}(x,y)$ as a function of $\tau'$. It returns -1 at $\tau' = 0$; it reaches its positive maximum (+1) at $\tau' = \tau'_{max,\tau}(x,y)$, and it is a piecewise constant function, so a "first positive step" value exists:

$$\tau'_{G,\tau}(x,y) = min\left(\tau' : \Psi^{\tau,\tau'}(x,y) > 0\right), \qquad (10)$$

which can be computed with the bisection method [10]. By means of the above described quantity, the set of "successful" pixels can be defined:

$$D_{G,\tau}^{\tau'} = \{(x,y) \in D : \Psi^{\tau,\tau'}(x,y) > 0\} = \{(x,y) \in D : \tau' \geq \tau'_{G,\tau}(x,y)\} \quad (11)$$

The relative amount of "successful" pixel with respect to the total area of the image can be regarded as a non decreasing function of the threshold $\tau'$:

$$\chi_\tau : [0, \tau'_{max,\tau}] \to [0,1] \quad (12)$$

$$\chi_\tau(\tau') = \frac{card(D_{G,\tau}^{\tau'})}{hw}. \quad (13)$$

Partitioning the range of such function into $k$ subintervals by means of $k+1$ values

$$\alpha'_j = \frac{j}{k} \quad j = 0,\ldots,k \quad (14)$$

corresponds to defining:

$$\tau'_{j,\tau} = min(\tau' : \chi_\tau(\tau') \geq \alpha_j), \quad (15)$$

The goal of tracing percentages of successful pixel to corresponding threshold values is attained by using the function

$$H' : \{\alpha'_j\} \to \{\tau'_{j,\tau}\} \quad (16)$$

$$H' : \{(\alpha'_j, \tau'_{j,\tau}); j = 0,\ldots,k\}. \quad (17)$$

**4 Setting the defect percentage in Directional Defect Map**
A slightly different procedure can be followed when dealing with the DDMAP. In this case, the amount of "defective" pixels referred to the total number of pixels in the image ranges from a minimum value of $0$ to a maximum which, generally speaking, is less than $1$. Let us use the symbol $\alpha^{\tau''}$ to identify the above mentioned ratio for a specific value $\tau''$ assigned to the DDMAP threshold:

$$\alpha^{\tau''} = \frac{card(D_{Y,\tau}^{\tau''})}{hw}, \quad (18)$$

where the set $D_{Y,\tau}^{\tau''}$ contains all defective points for that threshold level:

$$D_{Y,\tau}^{\tau''} = \{(x,y) \in D : \tilde{Q}^\tau(x,y) > (1+\tau'')\langle\tilde{Q}^\tau\rangle\}. \quad (19)$$

As $\tilde{Q}^\tau(x,y)$ takes limited values, then one can find a threshold value from which on the set $D_{Y,\tau}^{\tau''}$ is empty:

$$T'' = min(\tau'' \in R : D_{Y,\tau}^{\tau''} = \emptyset) \quad (20)$$

This means that the lower bound of $\alpha^{\tau''}$ is reached at this value of $\tau''$:

$$\min_{\tau'' \in [0, T'']} (\alpha_{\tau''}) = 0 = \alpha_{T''}. \tag{21}$$

The DDMAP algorithm suggests that the value of $T''$ can be computed as

$$T'' = \max_{(x,y) \in D} \left( \frac{\widetilde{Q}^\tau(x,y) - \langle \widetilde{Q}^\tau \rangle}{\langle \widetilde{Q}^\tau \rangle} \right). \tag{22}$$

On the other hand, the parameter $\tau''$ is greater or equal than zero and, $\alpha^{\tau''}$ is a non increasing function of $\tau''$, so that the maximum value of $\alpha^{\tau''}$ can be written as

$$\alpha^{\tau''}_{max} = \max_{\tau'' \in [0, T'']} (\alpha_{\tau''}) = \frac{\text{card}(D^0_{Y,\tau})}{hw}, \tag{23}$$

Now we can start from the extreme values of both $\tau''$ and $\alpha^{\tau''}$ listed above in order to find the intermediate values we need. Instead of considering a continuous range for these variables, which would imply a trivial subdivision of $[0, \alpha^{\tau''}_{max}]$ into $k$ subintervals and a consequent sequence of bisections, we must take into account the more practical case in which $\alpha^{\tau''}$ ranges from some units to some tenths percent, and we need a reasonable set of intermediate points, referring to values of $\alpha^{\tau''}$ represented by multiples of $0.01$. Following the same conceptual framework, a useful integer value corresponding to $\alpha^{\tau''}_{max}$ is represented by

$$\alpha_{max} = \lceil 100 \alpha^{\tau''}_{max} \rceil. \tag{24}$$

First of all, we set a parameter $k$, which corresponds to the number of *regularly spaced* values the interval of $\alpha^{\tau''}$ values can be divided in. The set value 100 is used to have the percentage from a simple ratio. The exact meaning of this parameter can be gathered from the following discussion. The remainder

$$r = \min(r' \in \mathbb{N} : \alpha_{max} \equiv r' (\text{mod} k)) \tag{25}$$

allows us to set the number of intermediate values for $\alpha^{\tau''}$:

$$\alpha''_j = j\delta; \quad j \in \{1, \ldots, j_{max}\} \subset \mathbb{N} \tag{26}$$

where

$$j_{max} = \begin{cases} k - 2 & (r = 0) \\ k - 1 & (r \neq 0). \end{cases} \tag{27}$$

and

$$\delta = \frac{\alpha_{max}}{100 j_{max}}. \tag{28}$$

For each value $\alpha_j$ the corresponding minimum threshold $\tau''_{j,\tau}$ can be found, for instance by means of a bisection algorithm, as described in the previous paragraph:

$$\tau''_{j,\tau} = min\left(\tau'' : \frac{\text{card}(D^{\tau''}_{Y,\tau})}{hw} \geq \alpha_j\right). \tag{29}$$

Thus, a function which can trace percentages of defective pixels to corresponding threshold values is

$$H'' : \{\alpha''_j\} \cup \{0, \alpha_{max}\} \rightarrow \{\tau''_{j,\tau}\} \cup \{0, T''\} \tag{30}$$

$$H'' : \{(\alpha''_j, \tau''_{j,\tau}); j = 1, \ldots, k\} \cup \{(0, T'')\} \cup \{(\alpha_{max}, 0)\}. \tag{31}$$

## 5 Examples

A tool has been developed to provide the calculation of CLD and the displaying of SMAP, DMAP and DDMAP. It works on any BMP or JPG picture. This package runs on Windows NT/2K with .NET, which is free at the MS site. The tool has an interface which is quite simple: it is possible to open a selected image and set the suitable processing parameters. It is can be download free at the following URL: http://staff.polito.it/roberto.marazzato/.

The first icon on the toolbar, showing a folder, allows to choose the image file (see the snapshot in Fig.7). The second icon allows the activation of the optimization procedures on CLD/SMAP and on DMAP/DDMAP. Let us consider the optimization procedure on the test image in Fig. 5. The optimized threshold on CLD and SMAP gives us a value of 45% and produces the DMAP and DDMAP shown in Fig.8. The consequent optimized thresholds for DMAP and DDMAP are 32% and 30% respectively.

We could ask ourselves about the possibility to specify the percentage of successful pixels, that is the green ones in the DMAP and the yellow ones in the DDMAP. After optimizing, we can decide the percentage of green and yellow pixels by choosing the third icon of toolbar: the tool provides the corresponding values of the thresholds. The example in Fig.9, with a high percentage of green pixels, allows to determine the position of what we can consider a defect of the texture, that is the lacking of a black square on the board. In fact, it is up to the analyst to decide under which circumstances the texture has a defect, and when it can be considered as regular.

An interesting use of CLD optimization is in applying it to segmentation procedures. In Fig.10.a, an image composed by two textures from Brodatz album [9] is subjected to the previously described procedure for optimizing the CLD threshold. After finding $\tau$, the optimized DMAP is determined: it turns out as the map in Fig.10.c. Note that the red pixels are concentrated in the region where the different texture is placed. Two other images (10.b and 10.d) are shown with different percentages of green pixels. Considering the dominant colour in specific areas, the DMAP corresponds to a segmentation procedure. Future investigation will be devoted to the used of DDMAP for segmentation.

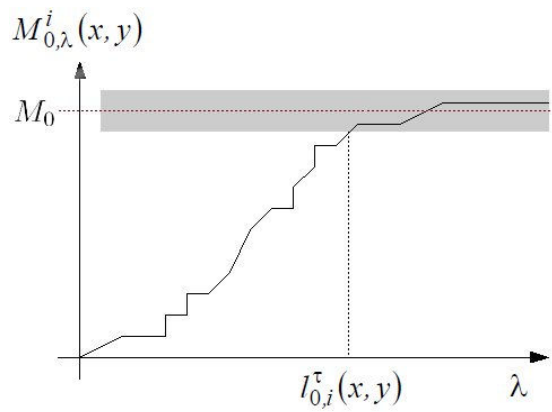

Figure 1: Saturation of 0-th order momentum.

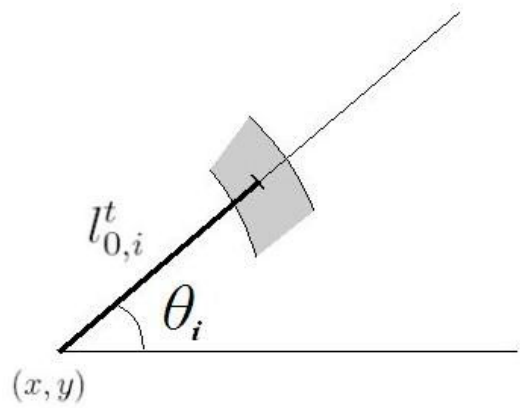

Figure 2: DMAP threshold, with shown a specific direction.

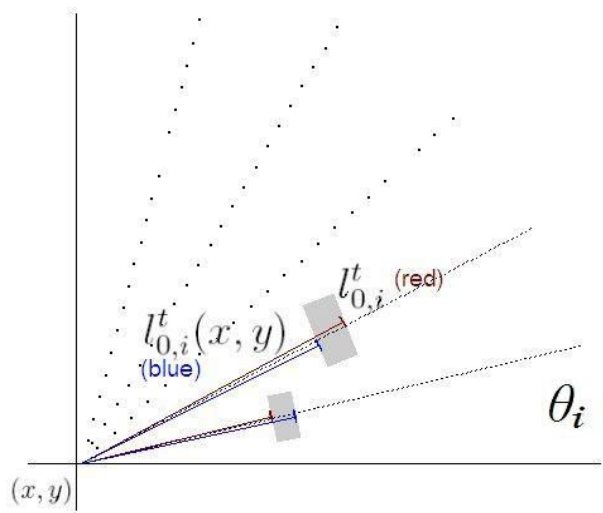

Figure 3: Comparison of local and overall CLDs.

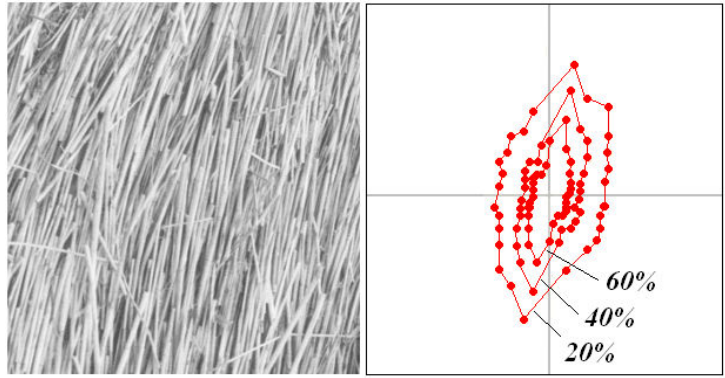

Figure 4: Behaviour of CLD for three different thresholds.

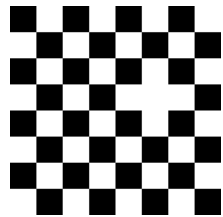

Figure 5: The sample image used in the following analysis.

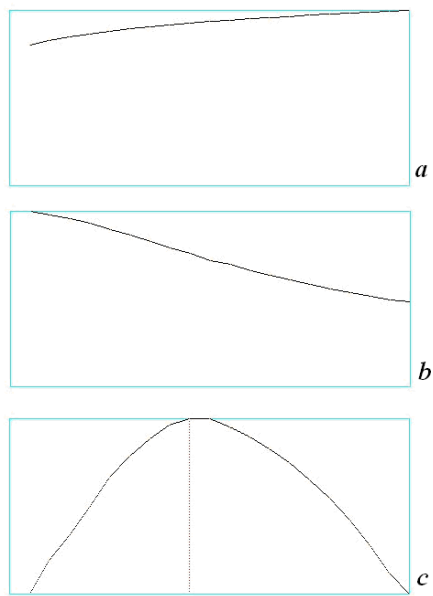

Figure 6: Behaviour of functions $\Omega(\tau)$ (a), $\omega(\tau)$ (b) and product $\Pi(\tau)$, which shows a maximum (c).

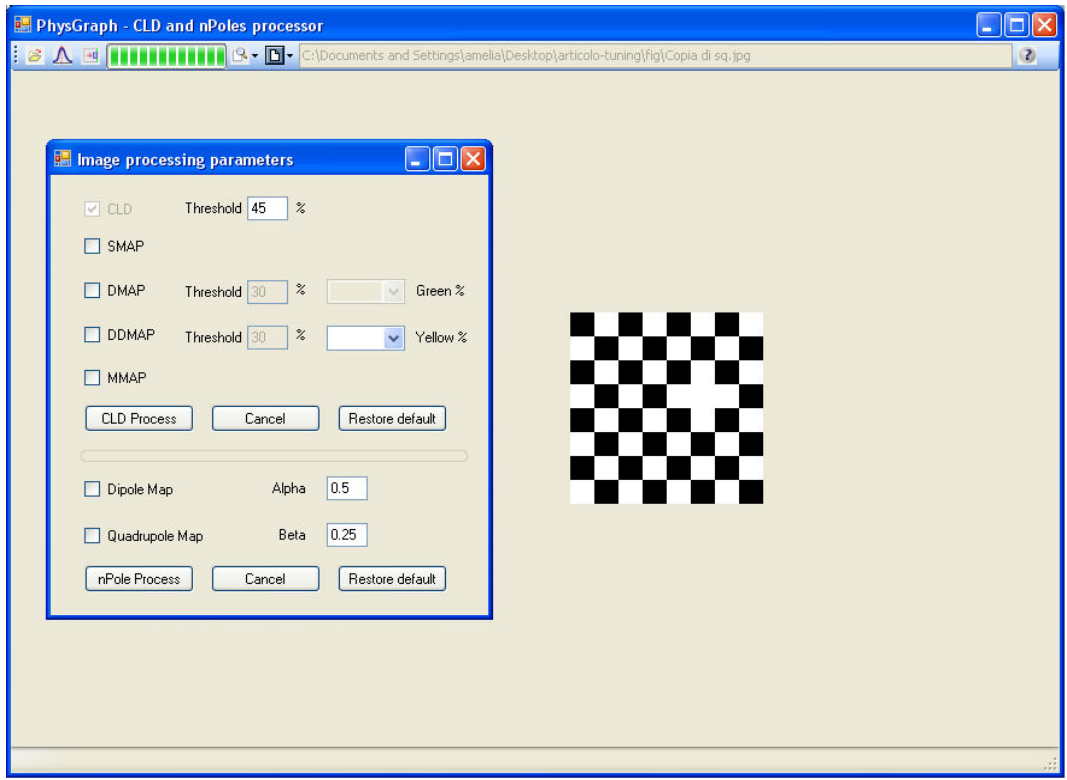

Figure 7: Screenshot of the CLD tool, showing the optimization of CLD threshold.

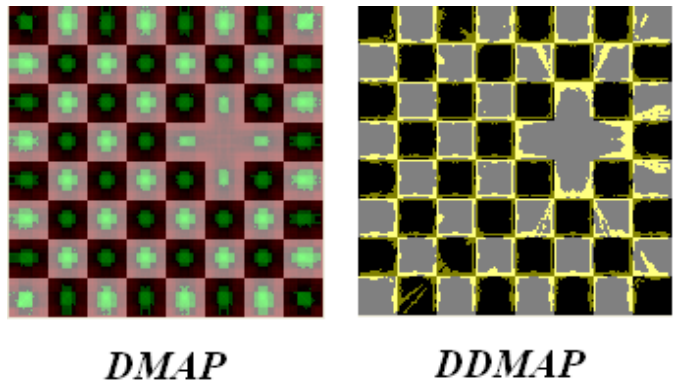

Figure 8: Optimized DMAP and DDMAP. The successful pixels in the DMAP are green.

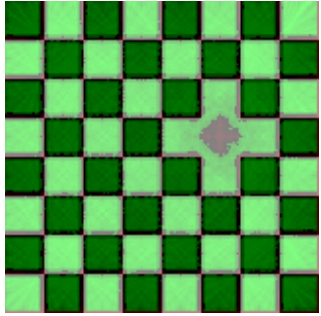

Figure 9: DMAP obtained by setting to 80% the percentage of successful (green) pixels.

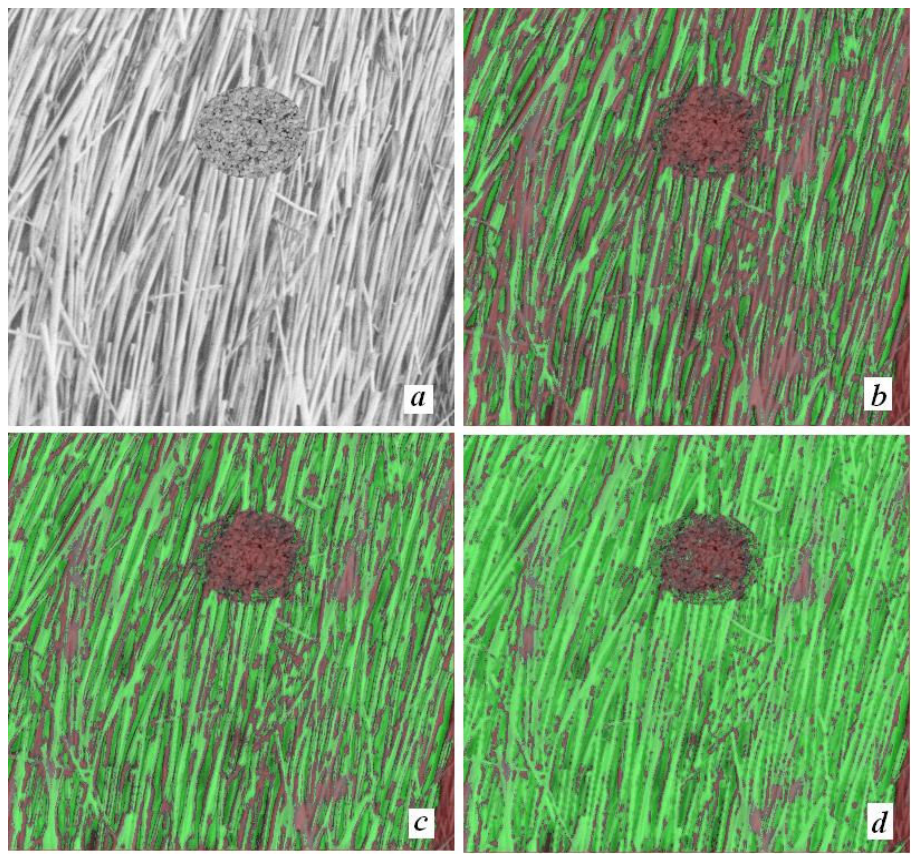

Figure 10: An example of segmentation through the CLD optimization. (a) is the image obtained from two texture of the Brodatz album. The optimization of CLD gives a threshold of 20%. Optimization of DMAP has a threshold of 83%, corresponding to a percentage of green pixels of 60%, as shows in (c). (b) corresponds to a covering of 40% and (d) of 80%.